\documentclass[conference]{IEEEtran}
\IEEEoverridecommandlockouts
\usepackage{caption}
\usepackage{cite}
\usepackage{amsmath,amssymb,amsfonts}
\usepackage{graphicx}
\usepackage{textcomp}
\usepackage{url}
\usepackage{xcolor}
\usepackage{amsthm}
\newtheorem{problem}{Problem}

\usepackage{algorithm}
\usepackage{algpseudocode}
\usepackage{booktabs}

\usepackage{listings}
\usepackage{xcolor}

\def\BibTeX{{\rm B\kern-.05em{\sc i\kern-.025em b}\kern-.08em
    T\kern-.1667em\lower.7ex\hbox{E}\kern-.125emX}}

\begin{document}

\title{Enabling Physical AI at the Edge: Hardware-Accelerated Recovery of System Dynamics
}

\author{\IEEEauthorblockN{Bin Xu}
\IEEEauthorblockA{\textit{School of Electrical Engineering} \\
\textit{Arizona State University}\\
Tempe, USA \\
binxu4@asu.edu}
\and
\IEEEauthorblockN{Ayan Banerjee}
\IEEEauthorblockA{\textit{School of Computing Science} \\
\textit{Arizona State University}\\
Tempe, USA \\
Ayan.Banerjee@asu.edu}
\and
\IEEEauthorblockN{Sandeep Gupta}
\IEEEauthorblockA{\textit{School of Computing Science} \\
\textit{Arizona State University}\\
Tempe, USA \\
Sandeep.Gupta@asu.edu}
}

\maketitle

\begin{abstract}
Physical AI at the edge—enabling autonomous systems to understand and predict real-world dynamics in real-time—demands efficient hardware acceleration. Model recovery (MR), which extracts governing equations from sensor data, is critical for safe and explainable monitoring in mission-critical autonomous systems (MCAS) operating under severe time, compute, and power constraints. While Field Programmable Gate Arrays (FPGAs) offer promising reconfigurable hardware for edge deployment, state-of-the-art (SOTA) MR methods like EMILY and PINN+SR rely on Neural ODEs requiring iterative solvers that resist hardware acceleration.
This paper presents MERINDA (Model Recovery in Dynamic Architecture), an FPGA-accelerated framework specifically designed to enable physical AI at the edge. MERINDA replaces computationally expensive Neural Ordinary Differential Equation (ODE) components with a hardware-friendly architecture combining: (a) Gated Recurrent Unit (GRU) layers for discretized dynamics, (b) dense inverse ODE layers, (c) sparsity-driven dropout, and (d) lightweight ODE solvers—with critical components fully parallelized on FPGA.
Evaluated on four benchmark nonlinear dynamical systems, MERINDA achieves transformative improvements over Graphics Processing Unit (GPU) implementations: \textbf{114× reduction in energy consumption} (434J vs. 49,375J), \textbf{28× smaller memory footprint} (214MB vs. 6,118MB), and \textbf{1.68× faster training}—while maintaining model recovery accuracy equivalent to SOTA methods. These results validate MERINDA's capability to bring physical AI to resource-constrained edge devices for real-time autonomous system monitoring.
\end{abstract}


\section{Introduction}
Physical AI\cite{dewi2025systematic} at the edge is revolutionizing autonomous systems by enabling real-time understanding of physical dynamics directly on resource-constrained devices. Unlike traditional cloud-based AI, edge deployment demands extreme efficiency in energy, memory, and latency—particularly critical for MCAS such as automated insulin delivery (AID), autonomous vehicles, and robotic systems where decisions must be made in milliseconds with minimal power consumption.

Data-driven inference for physical systems falls into two paradigms: (a) model learning (ML), which fits black-box neural networks to sensor data, and (b) model recovery (MR), which extracts interpretable governing equations—as exemplified by PiNODE~\cite{PiNode}, PINN+SR~\cite{chen2021physics}, and EMILY~\cite{pmlr-v255-banerjee24a,Banerjee2024,Xu2025MERINDA}. While ML offers predictive power, MR provides explainability and physical insight essential for safety-critical applications.

\begin{figure}
\centering\includegraphics[width=0.6\columnwidth,clip=false,trim=0 75 0 0]{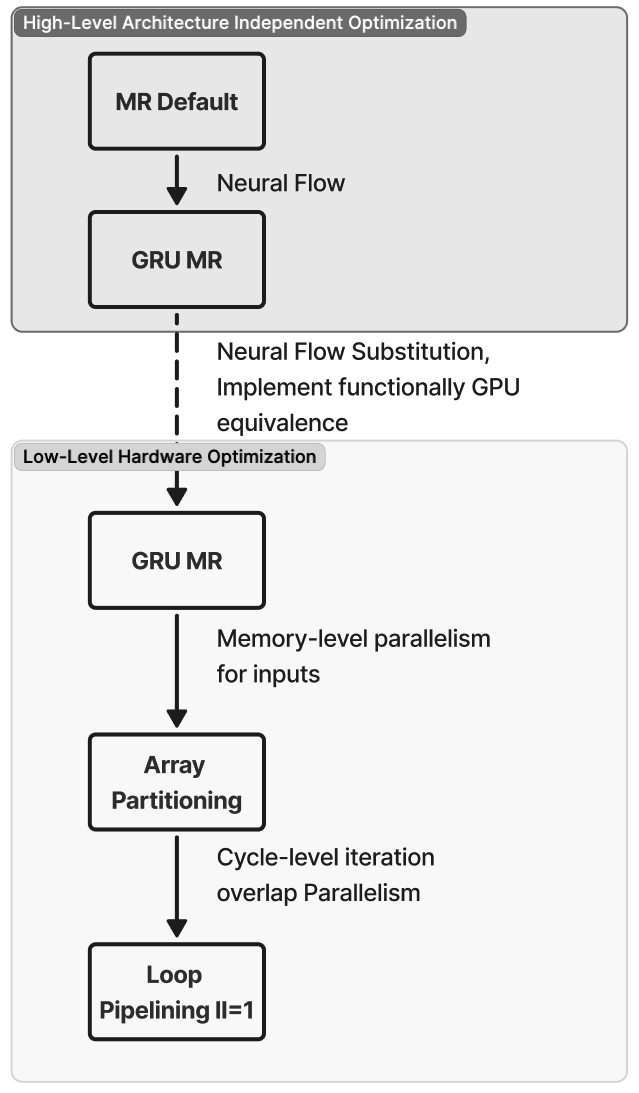}
    \vspace{0.2in}
    \caption{Optimization Framework of MERINDA. At the high level, we reformulate MR by replacing the iterative ODE solver in Liquid Time-Constant networks(LTC) with a GRU-based neural flow. At the low level, we specialize the GRU-based MR for FPGA by fine-grained spatial parallelism.}
    \label{fig:optimization_framework}
\end{figure}

\textbf{
The Edge AI\cite{gill2025edge,Xu2025DTLearning,Xu2025DigitalTwinning} Challenge:} Deploying MR at the edge faces a fundamental bottleneck. Current SOTA methods rely on Neural ODEs (NODEs), requiring iterative solvers that consume orders of magnitude more energy and time than standard neural networks. 
\noindent\textbf{Case Study: AID.}
Consider a wearable insulin pump for Type~1 Diabetes patients. The device must continuously learn personalized glucose--insulin dynamics to adapt to changing physiology, meals, and exercise, requiring model updates every five minutes. A GPU-based solution consuming 29{,}943.68~J per training cycle would require more than seven times the energy stored in a typical smartwatch battery (3.7\,V, 300\,mAh $\approx$ 4\,kJ), making even a single on-device update infeasible. In contrast, MERINDA requires only 261.79~J per update, enabling approximately fifteen full training cycles per battery charge. This shifts continuous on-device learning from impractical to deployable for medical wearables.

\begin{table}[h!]
\centering
\caption{MERINDA Performance Summary: Enabling Physical AI at the Edge (Hidden Size 16, Model Recovery Task)}
\scriptsize
\begin{tabular}{lccc}
\toprule
\textbf{Metric} & \textbf{GPU} & \textbf{FPGA (MERINDA)} & \textbf{Speedup} \\
\midrule
Training Time (s) & 163.51 & 55.23 & \textbf{2.96$\times$ faster} \\
Energy (J)        & 29,943.68 & 261.79 & \textbf{114$\times$ less} \\
DRAM (MB)         & 5,862.32 & 211.29 & \textbf{28$\times$ smaller} \\
Accuracy (MSE)    & 3.18 & 5.37 & Comparable \\
\bottomrule
\end{tabular}
\label{tab:summary_impact}
\end{table}

\begin{figure}[h]
\centering\includegraphics[width=0.8\columnwidth,clip=false,trim=0 75 0 0]{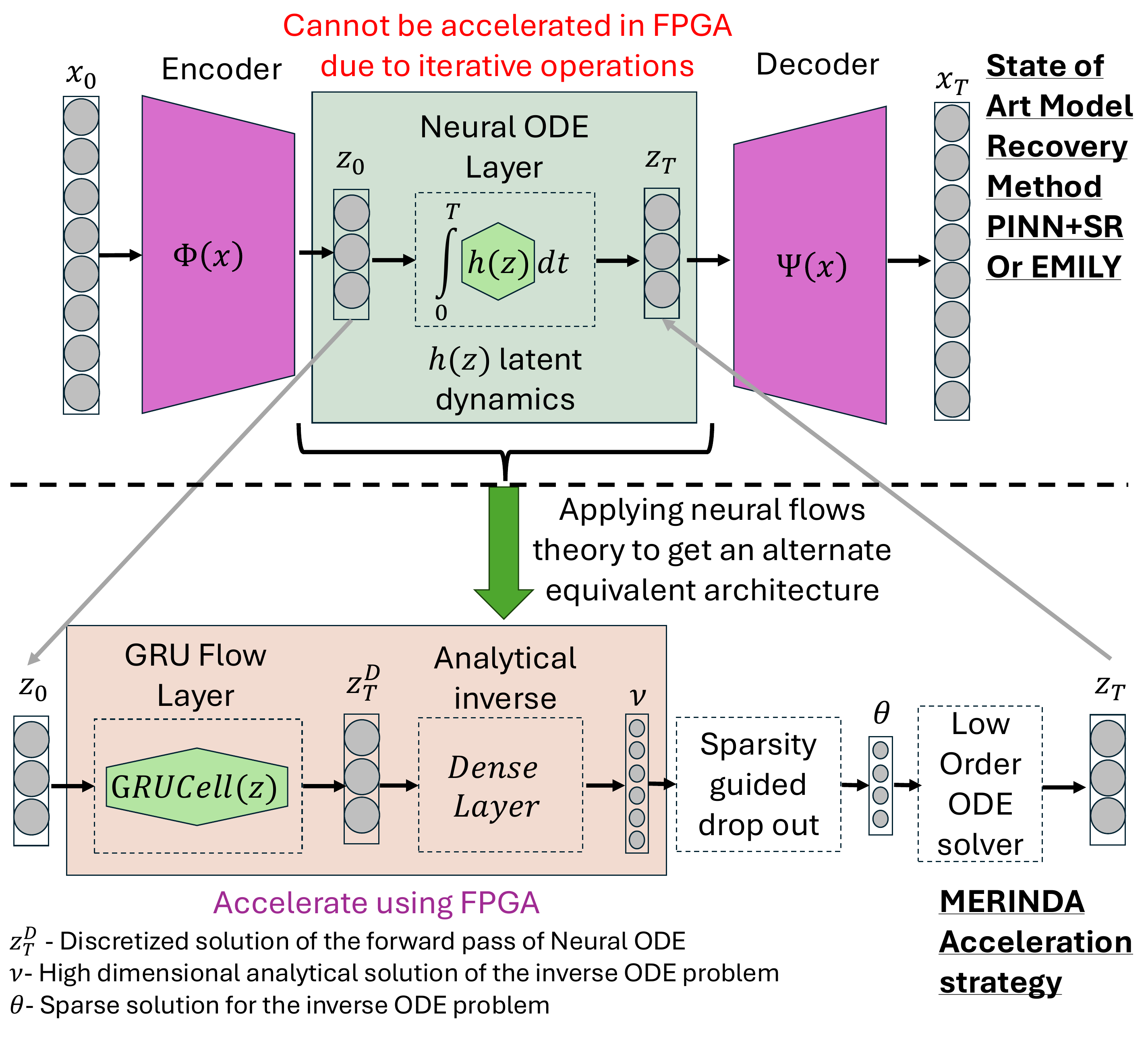}
    \caption{FPGA acceleration strategy using neural flow based equivalent architecture to neural ODEs.}
    \label{fig:FPGA} 
\end{figure}

\textbf{Hardware Acceleration Challeng:}
Traditional time-series ML accelerators succeed because of structured, discretized Recurrent Neural Network (RNN) operations implementable via matrix multiplications and specialized units like CORDIC~\cite{Cao20Cordic}. However, SOTA MR methods (EMILY, PINN+SR, PiNODE) rely on Neural ODE layers requiring \textit{iterative solvers} that resist parallelization (Fig.~\ref{fig:FPGA}). Prior ODENet acceleration efforts~\cite{Wtanabe,CaiFPGA} assume fixed architectures, while real MR demands adaptive NODE depth. Existing ODE solver accelerators~\cite{stamoulias2017high,ebrahimi2017evaluation} assume static coefficients, incompatible with learning-based methods where ODEs change during training.

FPGAs offer a promising solution through reconfigurable, energy-efficient hardware acceleration. While FPGA-based ML acceleration has been extensively studied~\cite{el2024fpga,ma2018optimizing,FPGATrans}, MR acceleration remains largely unexplored due to the difficulty of parallelizing adaptive ODE solvers. This gap prevents physical AI from achieving its full potential at the edge, where recent trends in Edge AI and on-device intelligence~\cite{gunter2024apple} demonstrate growing demand.

This paper presents MERINDA, a novel framework (Fig.~\ref{fig:optimization_framework}) that makes physical AI practical at the edge through FPGA acceleration. Leveraging neural flow theory~\cite{bilovs2021neural} which is shown in Fig.~\ref{fig:FPGA}, we replace computationally expensive NODE layers with hardware-friendly invertible functions using GRU and dense neural layers—achieving theoretical equivalence while enabling massive parallelization on reconfigurable hardware.

Our work demonstrates that physical AI can operate efficiently on edge devices with dramatic improvements, which is shown in Tab.~\ref{tab:summary_impact}.
\noindent{Key Contributions:}

\noindent{\bf 1. Hardware-Accelerated Physical AI Framework (Fig. \ref{fig:optimization_framework}):} We present the first FPGA-accelerated model recovery architecture achieving GPU-equivalent accuracy with order-of-magnitude improvements in energy efficiency and memory footprint—validated on four benchmark nonlinear systems including real-world automated insulin delivery.

\noindent{\bf 2. Comprehensive Performance Analysis:} We compare MERINDA against three paradigms (ML, physics-guided ML, and full MR) on both FPGA and GPU platforms, revealing fundamental trade-offs between accuracy, latency, energy, and memory—providing practitioners with actionable deployment guidance.

\section{Theoretical background}
This section presents basics of MR and establishes approximate equivalence of neural flow architecture with NODE.
\subsection{Basics of Model Recovery}
The main goal of MR is akin to an auto-encoder (Figure \ref{fig:FPGA}), where given a multivariate time series signal $X(t)$, the aim is to find a latent space representation that can be used to reconstruct an estimation $\Tilde{X}(t)$ with low error. It has the traditional encoder $\phi(t)$ and decoder ($\Psi(t)$) of an autoencoder architecture. MR represents the measurements $X$ of dimension $n$ and $N$ samples, as a set of nonlinear ordinary differential equation model in Eqn: \ref{eqn:Model}. 
\begin{equation}
    \label{eqn:Model}
    \scriptsize
    \dot{X} = h(X,U,\theta),
\end{equation}
\vspace{-0.2in}\\
where $h$ is a parameterized non-linear function, $U$ is the $m$ dimensional external input, and $\theta$ is the $p$ dimensional coefficient set of the nonlinear ODE model.

\noindent{\bf Sparsity:} An $n$-dimensional model with $M^{th}$ order non-linearity can utilize $\binom{M+n}{n}$ non-linear terms. A sparse model only includes a few non-linear terms $p << \binom{M+n}{n}$. Sparsity structure of a model is the set of non-linear terms used by it.

\noindent{\bf Identifiable model:} A model in Eqn.\  \ref{eqn:Model} is identifiable~\cite{verdiere2019systematic}, if $\exists$ time $t_I > 0$, such that $\forall \theta, \Tilde{\theta} \in \mathcal{R}^p$:
\begin{equation}
    \label{eqn:Ident}
    \scriptsize
    \forall t \in [0,t_I], f(X(t),U(t),\theta) = f(X(t),U(t),\Tilde{\theta}) \implies \theta = \Tilde{\theta}. 
\end{equation}

\noindent Eqn. \ref{eqn:Ident} effectively means that a model is identifiable if two different model coefficients do not result in identical measurement $X$. In simpler terms this means $\forall \theta_i \in \theta, \frac{dX}{d\theta_i} \neq 0$. In this paper, we assume that the underlying model is identifiable.

\begin{problem}[Sparse Model Recovery]\label{prob:Problem} Given $N$ samples of measurements $X$ and inputs $U$, obtained from a sparse model in Eqn. \ref{eqn:Model} such that $\theta$ is identifiable, recover $\Tilde{\theta}$ such that for $\Tilde{X}$ generated from $f(X,U,\Tilde{\theta})$, we have $||X - \Tilde{X}|| \leq \epsilon$, where $\epsilon$ is the maximum tolerable error.
\end{problem}

\noindent{\bf Role of NODE:} Both EMILY~\cite{pmlr-v255-banerjee24a} and PINN+SR~\cite{chen2021physics} utilize a layer of NODE cells in order to integrate the underlying non-linear ODE dynamics. NODE cell's forward pass is by design the integration of the function $h$ over time horizon $T$ with $N$ samples (Fig. \ref{fig:FPGA}). This effectively requires an ODE solver in each cell of the NODE layer:
\vspace{-0.1in}\\
\begin{equation}
\scriptsize
z(t) = \int\limits^T_0{h(z,u,\theta)dt}, 
\end{equation}
\vspace{-0.15in}\\
where $z \in Z$ and $u \in U$ are each cells output and input. 

The results are then used further in the EMILY or PINN+SR pipeline to extract the accurate underlying non-linear ODE model.
\subsection{Neural flows and equivalent architectures to NODE}

According to the theory of neural flows~\cite{bilovs2021neural}, the node layer can be replaced by an approximate solution to $F(t) \approx Z(t)$ in discretized form using recurrent nerual network architectures such as GRU provided that following conditions are satisfied:

\begin{scriptsize}
\begin{equation}
    F(0,u) = Z(0,u), \text{(initial condition), and  }
    F(t,u) \text{  is invertible}.
\end{equation}
\end{scriptsize}

Bilovs et al.~\cite{bilovs2021neural} show that $F(t,u)$ can be achieved by replacing the original NODE layer by a GRU layer. However, the GRU layer does not statisfy by the invertible condition. The authors in~\cite{bilovs2021neural} suggest the usage of a dense layer since it acts as a universal approximator of nonlinear functions and hence can also act as the inversion of the function $F(t,u)$.

MERINDA further enhances the equivalent architecture proposed in~\cite{bilovs2021neural} by further pruning the dense layer as shown in Figure \ref{fig:Approach}. The main idea is to further reduce the dense layer structure by utilizing the inherent sparsity in the data.

Given the definitions of identifiability and sparsity, we now discuss our full architecture that is equivalent to PiNODE


\section{MERINDA Architecture}

\begin{figure}[h]
\centering
\includegraphics[width=\columnwidth]{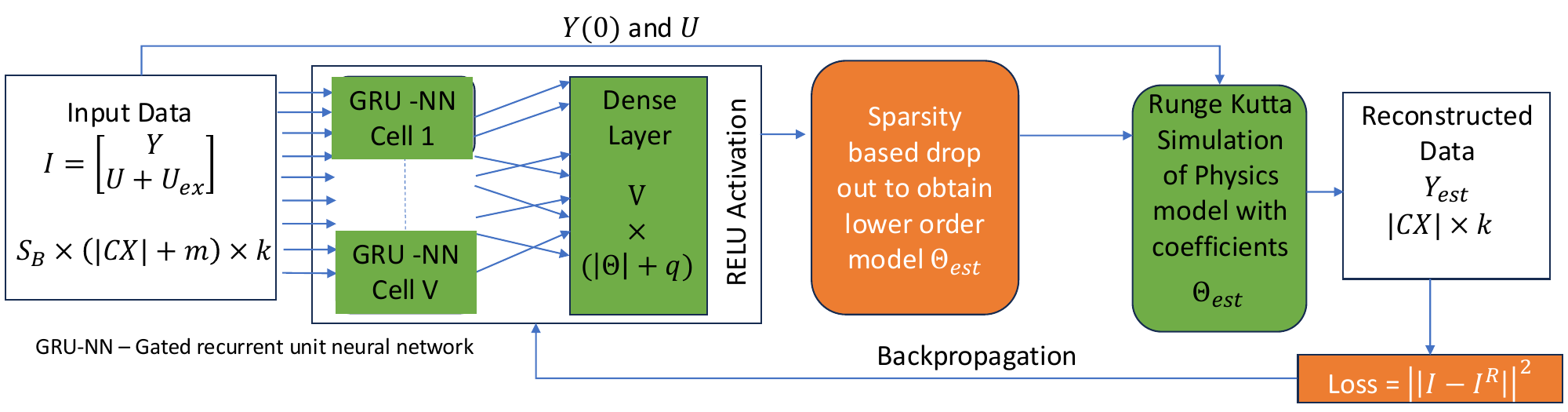} 
 \caption{MERINDA: Gated recurrent unit (GRU) NN-based MR architecture.} 
    \label{fig:Approach}
\end{figure}

In our approach (Fig. \ref{fig:Approach}), we extend gated recurrent unit neural network (GRU-NN) to obtain advanced neural structure MERINDA that can solve the model recovery problem. The forward pass of GRU-NN structure expresses the coefficients of the model as a nonlinear function of the outputs $Y$ and inputs $U$ of the model. The measurements of $Y$, can be used to convert the set of implicit dynamics to an overdetermined system of equations that are nonlinear in terms of the model coefficients. As such an over-determined system of equation may have no solution unless either some equations are rejected or are expressed as linear superposition of other equations. To search for a set of consistent equations to estimate model coefficient, a dense layer is utilized. The search process of the dense layer is guided by a loss function (\textit{ODE loss}) that computes the mean square error between the estimated $Y_{est}$ using an Ordinary Differential Equation (ODE) solver $\mathbf{SOLVE}(Y(0),\Theta, U)$ and the ground truth measurements $Y$.

The advanced neural architectures for model recovery in Fig. \ref{fig:Approach} is implemented by extending the base code available in~\cite{liquid-time-constant-networks}. We extract the training data consisting of temporal traces of $Y$, and $U$ . $Y$ is sampled at least at the Nyquist rate for the application, and $U$ has the same sampling rate as $Y$. The resulting training data is then divided into batches of size $S_B$. This forms a 3 D tensor of size $S_B \times |Y|+m \times k$. 

Each batch is passed through the {\it GRU-NN} network with $V$ nodes, resulting in $V$ hidden states. A dense layer is then employed to transform these $V$ hidden states into $p = |\Theta|$ model coefficient estimates and $q$ input shift values. The dense layer is a multi-layer perceptron with ReLU activation function for the model coefficient estimate nodes, whose outputs are the estimated model coefficients. The dense layer converts the $V$ dimensional hidden layer outputs to $M+|X| \choose |X|$ which is the number of non-linear terms that can be used for an $M$ th order polynomial. A dropout rate of $|\Theta|$ is used so that the final number of output layers with non-zero activation is $|\Theta|$. The model coefficient estimates and the initial value $Y(0)$ is passed through an ODE solver to solve the nonlinear dynamical equations with the coefficients $\Theta_{est}$, initial conditions $Y(0)$ and inputs $U$. The Runge Kutta integration method is used in the ODE solver, which gives $Y_{est}$. In the backpropagation phase the network loss is appended with ODE loss, which is the mean square error between the original trace $Y$ and estimated trace $Y_{est}$.

\subsection{Accelerator Design Exploration}
 



\subsubsection{Computation Optimization}
To achieve high-throughput GRU computation on hardware, we use a deeply pipelined architecture that allows a new GRU cell to start every clock cycle. Applying \texttt{\#pragma HLS PIPELINE II=1} on the outer loop over time ensures an initiation interval (II) of 1 for the entire GRU cell.

Inside the GRU cell, each stage—such as computing the reset gate, update gate, and the new hidden state—is decomposed into smaller element-wise operations instead of one large matrix--vector computation. This exposes fine-grained parallelism. We apply \texttt{\#pragma HLS UNROLL} to the inner loops so that operations across the input and hidden dimensions run fully in parallel, effectively forming parallel multiply--accumulate (MAC) arrays.

To further improve throughput, we use \texttt{\#pragma HLS DATAFLOW} so that independent stages execute concurrently and communicate through streaming FIFOs. This overlaps work between different compute stages and across successive time steps.
By combining dataflow pipelining, loop pipelining, and loop unrolling, the GRU design becomes latency-insensitive and reaches a throughput determined solely by the optimized initiation interval of 1.

\subsubsection{Memory Optimization}
To reduce off-chip memory access and improve data locality, we place frequently used variables in on-chip buffers using HLS pragmas. The hidden state buffer, which is reused across timesteps, is fully partitioned with \texttt{\#pragma HLS ARRAY\_PARTITION complete} to allow parallel access to all hidden units.

Temporary values used within each GRU computation step are selectively mapped to either registers or BRAM which is on-chip memory on FPGA based on their access patterns and reuse needs. Frequently accessed data is stored in registers for low-latency access, while larger or less frequently accessed data is placed in BRAM to balance storage efficiency and routing complexity.

The updated hidden state is directly forwarded to subsequent layers without sending it back to off-chip memory, further reducing bandwidth usage. Combined with loop pipelining and dataflow execution, these memory optimizations help maintain a one-cycle initiation interval (II=1), enabling high-throughput FPGA execution.


\subsection{MILP-Based Platform and Task Optimization Strategy}

In this paper, we propose a Mixed-Integer Linear Programming (MILP) approach to identify the optimal platform (FPGA or GPU), task type (ML, ML-PG, or MR), and corresponding hyperparameter configurations. The motivation for applying MILP to MERINDA is to systematically explore a broader design space and determine configurations that best align with real-world deployment constraints—such as energy, latency, and accuracy—enabling resource-aware model recovery in mission-critical settings.
For the equations below, $\text{Hi}$ represents the hidden layer size, $E$ denotes the number of epochs, $\text{sq}$ is the sequence length (input size), $P$ represents power, $M$ denotes memory, $\varepsilon$ is the error and $R$ is the threshold of time. We set the objective to find out the trade-off between power and memory with constraints in error $f_e(a, h, hi, e, N)$ and time $t_{\text{ex}} (a, h, hi, e, N)$. We employ Ridge regression strategy~\cite{mcdonald2009ridge} on polynomial library functions to model the error $f_e(a, h, hi, e, N)$, execution time $t_{\text{ex}}(a, h, hi, e, N)$, power $f_p(hi, e, N)$, and memory $f_m(hi, e, N)$ which are cubic functions. \

\begin{figure}[h!]
    \centering
    \scriptsize
    \begin{tabular}{ll}
        \textbf{Input:} & $\vec{X} = 
        \begin{bmatrix}
        x_1 & x_2 & \cdots & x_n
        \end{bmatrix}$ \\[1ex]
        \textbf{Platforms:} & $a \in A = \{0, 1\}$ \quad $ (0 \rightarrow \text{FPGA}, \, 1 \rightarrow \text{GPU})$ \\[1ex]
        \textbf{Tasks:} & $h \in H = \{0, 1, 2\}$ \quad $(0 \rightarrow \text{ML}, \, 1 \rightarrow \text{ML + PG}, \, 2 \rightarrow \text{MR})$ \\[1ex]
        \textbf{Hyper:} & $hi \in \text{Hi} = \{16, 32, 64, 128\}, \,  e \in E = \{16, 32, 64, 128\},$ \\[1ex]
        & $N \in \text{sq} = \{50, 100, 200\}$ \\[1ex]
        \textbf{Objective:} & $O = \gamma P + (1 - \gamma) M, \, P = f_p(hi, e, N),$ \\[1ex]
        & $M = f_m(hi, e, N)$ \\[1ex]
        \textbf{Goal:} & $\left\lVert \hat{x}t - xt \right\rVert \leq \varepsilon$ \\[1ex]
        \textbf{Constraints:} & $f_e(a, h, hi, e, N) < \varepsilon, \quad t_{\text{ex}} (a, h, hi, e, N) < R$ \\[1ex]
        \textbf{Optimization:} & $a, h, hi, e, N \quad \text{to minimize } O = \gamma P + (1 - \gamma) M$
    \end{tabular}  
    \caption{Mixed Integer Linear Program to obtain optimal approach and hyperparameters for MR, ML or ML-PG.}
    \label{fig:problem_formulation}
\end{figure}

To perform optimization for the continuous variable in our MILP setup, we utilize the L-BFGS-B algorithm. L-BFGS-B stands for Limited-memory Broyden Fletcher Goldfarb Shanno with Box constraints, an optimization algorithm that is particularly well-suited for high-dimensional problems where memory efficiency is crucial\cite{zhu1997algorithm}. We employ L-BFGS-B with the average error as a continuous variable, while keeping the other hyperparameters fixed at discrete values.

\section{Evaluation and Results}
\subsection{Data Description}\label{AA}
\noindent{\bf Simulation case studies:} All data was generated by implementing the models in Matlab and using ODE 45 to solve the ODEs. 200 s of each model was generated at Nyquist sampling rate to obtain the data for the simulative case studies. 

\noindent{\bf Real world case studies:} Each case study had their own real world datasets.

\noindent\underline{Lotka Volterra:} We used yearly lynx and hare pelts data collected from Hudson Bay Company~\cite{kaiser2018sparse}.

\noindent\underline{Pathogenic attack system: } The data is available in~\cite{kaiser2018sparse}.

\noindent\underline{AID:} The datasets are virtual patient data obtained from the OhioT1D dataset available in~\cite{marling2020ohiot1dm}. It is 14 time series data of glucose insulin dynamics. Each time series data had a duration of 16 hrs 40 mins which amounts to 200 samples of CGM and insulin data. In each time series, meal ingestion time was varied from $t = 15$ min to $t = 400$ min with carbohydrate value in the range $[0g, 28g]$ for each meal, and bolus insulin delivery in the range $[0 U, 40 U]$. 

\subsection{Implementation Details}\label{AA}
To evaluate the performance of the FPGA, we perform experiments on GPU and FPGA with the same dataset. The performance of GPU is set as baseline. 

\paragraph{GPU Platform}
The experiments were conducted using an Intel Xeon w9-3475X CPU and an NVIDIA RTX 6000 GPU with 48GB of memory. The model was loaded using TensorFlow 2.10 and Keras 2.10. We monitor power consumption using {\tt nvidia-smi} and measure execution time and DRAM footprint using the {\tt time} and {\tt psutil} library.


\paragraph{FPGA Platform}
For the FPGA platform, the experiments were performed on a PYNQ-Z2 board with a Dual-Core ARM Cortex-A9 processor and a 1.3M-configurable-gate FPGA. The GRU model was built from scratch, with the forward pass and backpropagation logic developed in C++ using High-Level Synthesis (HLS) on AMD’s Vitis tool. Afterward, in Vivado, the forward pass accelerator was integrated with the FPGA board using Direct Memory Access (DMA) to interface with the processing system on the PYNQ-Z2. We monitor the power of hardware using Vivado power report and measure time and DRAM footprint using {\tt time} and {\tt psutil} libraries.


\paragraph{Applying Mixed-Integer Linear Programming}
We employ MILP to determine the optimal hyperparameters for different platform-task combinations and to find an trade-off between DRAM footprint and energy consumption.  
\noindent\textbf{Functions expressing time, memory, power and error in terms of hyper-parameters:} From the experimental results, we performed Ridge regression to obtain function \eqref{eqn:13} of DRAM footprint with ML-PG. In the function \eqref{eqn:13} \text{e} is epoch, \text{N} is sequence length, \text{hi} is hidden layer size and $\boldsymbol{\varepsilon}$ is error.  
\begin{equation}
\scriptsize
\text{DRAM\_access} = 4.7316 \cdot \text{e} - 194.3639 \cdot \text{N} + 39.4598 \cdot \text{hi} - 2.4789 \cdot \varepsilon + 503.8408
\label{eqn:13}
\end{equation}
We also have another 17 functions, including the DRAM footprint, energy consumption, and time with ML, ML-PG and MR respectively. In the interest of time we do not show them in this paper.

\subsection{Results}\label{AA}



    
    


\subsubsection{Accuracy comparison of MERINDA with EMILY and PINN+SR} We compare the accuracy of MERINDA using the MSE metric. Table \ref{tbl:SMRFM} shows the errors of MERINDA, PINN+SR and EMILY on the benchmark examples available in~\cite{kaiser2018sparse}. EMILY and PINN+SR's accuracy results are directly obtained from~\cite{pmlr-v255-banerjee24a} and \cite{robinson2022physics}. 

\begin{table}[h!]
\centering
\scriptsize
\caption{Comparison between MERINDA and SOTA MR techniques EMILY and PINN+SR using reconstruction MSE. Errors are absolute values; numbers in parentheses indicate standard deviation.}
\begin{tabular}{lccc}
\toprule
\textbf{Example} & \textbf{EMILY} & \textbf{PINN+SR} & \textbf{MERINDA} \\
\midrule
Lotka Volterra      & 0.03 (0.02)     & 0.05 (0.03)     & 0.03 (0.018) \\
Chaotic Lorenz      & 1.7 (0.6)       & 2.11 (1.4)      & 1.68 (0.4) \\
F8 Cruiser          & 4.2 (2.1)       & 6.9 (4.4)       & 5.1 (2.2) \\
Pathogenic Attack   & 14.3 (12.1)     & 21.4 (5.4)      & 15.1 (10.2) \\
\bottomrule
\end{tabular}
\label{tbl:SMRFM}
\end{table}

\begin{table*}[h]
\centering
\scriptsize
\caption{performance comparison across FPGA, Mobile GPU, and GPU implementations for ML, ML+PG, and MR tasks across different hidden sizes.}
\renewcommand{\arraystretch}{1.1}
\resizebox{\textwidth}{!}{%
\begin{tabular}{lccccccc}
\toprule
\textbf{Metric} & \textbf{Size} & \textbf{FPGA (ML)} & \textbf{GPU (ML)} & \textbf{FPGA (ML + PG)} & \textbf{GPU (ML + PG)} & \textbf{FPGA (MR)} & \textbf{GPU (MR)} \\
\midrule
\textbf{Average Error} & 16 & 7.33 & 8.5603 & 6.79 & 7.309 & 5.3678 & 3.179 \\
                       & 32 & 7.872 & 7.2366 & 8.29 & 7.257 & 4.91 & 3.54 \\
                       & 64 & 7.769 & 7.772 & 6.478 & 7.762 & 5.77 & 3.1157 \\
                       & 128 & 7.37 & 7.5498 & 8.02 & 6.931 & 4.6 & 3.2965 \\
\midrule
\textbf{Time of Training (s)} & 16 & 37.37 & 30.83 & 32.2 & 30.31 & 55.23 & 163.51 \\
                              & 32 & 39.62 & 29.98 & 33.36 & 29.21 & 54.7 & 145.87 \\
                              & 64 & 42.16 & 29.78 & 42.04 & 30.45 & 63.69 & 152.55 \\
                              & 128 & 66.74 & 29.41 & 66.74 & 29.78 & 88.5 & 149.14 \\
\midrule
\textbf{Energy Consumption (J)} & 16 & 177.13 & 5457.24 & 152.63 & 5488.00 & 261.79 & 29943.68 \\
                                & 32 & 193.23 & 5438.65 & 162.70 & 5437.39 & 266.77 & 27403.20 \\
                                & 64 & 208.52 & 5477.22 & 207.93 & 5459.41 & 315.01 & 29206.32 \\
                                & 128 & 327.36 & 5560.32 & 327.36 & 6236.72 & 434.09 & 27375.12 \\
\midrule
\textbf{DRAM Footprint (MB)} & 16 & 210.46 & 4399.70 & 225.07 & 4344.71 & 211.29 & 5862.32 \\
                             & 32 & 210.94 & 4399.61 & 225.11 & 4351.26 & 210.63 & 5881.50 \\
                             & 64 & 209.55 & 4404.27 & 226.61 & 4363.09 & 211.72 & 5947.67 \\
                             & 128 & 214.45 & 4433.51 & 229.14 & 4377.54 & 214.23 & 6118.36 \\
\bottomrule
\end{tabular}
}
\label{tab:fpga_gpu_comparison_all}
\end{table*}

\subsubsection{Support for varying model coefficients}
In the MERINDA architecture, at each forward pass we solve a set of ODE, which is setup by the dense layer outputs. The output of each neuron in the dense layer corresponds to the coefficients of a non-linear component. If we choose a polynomial order of M, then the number of neurons in the dense layer should be $M+n \choose n$ for each of the variables. Initially the dense layer outputs are random based on the initial weights. This results in the first set of differential equations. For example, when $M = 2$ and $n = 2$ for the LOTKA Volterra system, then we should have $12$ neurons for two equations in the dense layer with random weights resulting in random outputs: $[0.4 0.5 0.6 0.1 0.2 0.5 0.1 0.3 0.4 0.6 0.8 0.2]$. This results in an ODE for the first forward pass as shown in the equation below:

\begin{scriptsize}
\begin{eqnarray}
\label{eqn:Eq1}
\dot{x_1} = 0.4u + 0.5 x_1 + 0.6 x_2 + 0.1x_1^2 + 0.2 x_2^2 + 0.5 x_1x_2,\\\nonumber 
\dot{x_2} = 0.1 u + 0.3 x_1 + 0.4 x_2 + 0.6 x_1^2 + 0.8 x_2^2 + 0.2 x_1x_2
\end{eqnarray}
\end{scriptsize}
We then apply a threshold based drop out. For the lotka Volterra model the threshold was set at 0.001. With this threshold none of the dense layer nodes are dropped out. Hence Eqn. \ref{eqn:Eq1} becomes the first ODE that we need to solve in the first forward pass. This equation is different from the LOTKA Volterra model. The weights of all the layers are then modified using backpropagation to minimize the loss between the data obtained from the Lotka Volterra model and the solution of the ODE in Eqn \ref{eqn:Eq1}. In the next iteration, we get an updated weight with the output of the dense layer as follows $[0.0006 0.55 0.06 0.0003 0.005 -0.09 0.8 0.003 -0.7 0.04 0.06 0.00005 0.008]$. After dropout the dense layer output becomes - $[0 0.55 0.06 0 0.005 -0.09 0.8 0.003 -0.7 0.04 0.06 0 0.008]$. This results in the following equation

\begin{scriptsize}
\begin{eqnarray}
\label{eqn:Eq2}
\dot{x_1} = 0.55 x_1 + 0.06 x_2 + 0.005x_1^2 – 0.09 x_2^2 + 0.8 x_1x_2 \\\nonumber
\dot{x2} = 0.003 u - 0.7 x_1 + 0.04 x_2 + 0.06 x_1^2  + 0.008 x_1x_2
\end{eqnarray}
\end{scriptsize}
This becomes the next equation that is solved in the next forward pass. This continues for 200 epochs and finally we arrive at the following output of the dense layer $[0 0.52 0 0 0 -0.026 0.999 0 -0.501 0 0 0.005]$. This means that the final ODE solved is as follows: 

\begin{scriptsize}
\begin{eqnarray}
\dot{x_1} = 0.52 x_1 - 0.026 x_1x_2 \\\nonumber
\dot{x_2} = 0.999 u - 0.501 x_2 + 0.005 x_1 x_2
\end{eqnarray}
\end{scriptsize}

\subsubsection{Performance comparison between GPU and FPGA}
As observed in Table~\ref{tab:fpga_gpu_comparison_all}, the FPGA implementation of MR shows substantial gains in energy efficiency, memory usage, and training time relative to the GPU counterpart. At hidden size 128, FPGA achieves a 1.68$\times$ speedup (88.5s vs. 149.14s) in training time. The improvement is more prominent in energy consumption, where FPGA consumes up to 114$\times$ less energy than GPU (434.09J vs. 49,375.12J at size 128). In terms of memory, FPGA exhibits over 28$\times$ lower DRAM usage at the same size (214.23MB vs. 6118.36MB), affirming its suitability for memory-constrained edge deployments. While GPUs maintain higher MR accuracy across sizes (e.g., 3.2965 vs. 4.6 at size 128), the trade-offs favor FPGA for applications prioritizing low power and resource efficiency with tolerable accuracy degradation, making it a compelling option for real-time MR in MCAS settings.
Table \ref{tab:fpga_resource_usage} summarizes the resource utilization for different matrix sizes. As observed, the accelerator performing matrix multiplication with a size of 128 exhibits high resource utilization. The computational resources are fully leveraged across larger matrix sizes.

\begin{table}[thbp]
\centering
\scriptsize
\caption{FPGA resource utilization (\%) across matrix sizes for key hardware components.}
\setlength{\tabcolsep}{4pt}
\renewcommand{\arraystretch}{0.95}
\begin{tabular}{lccccc}
\toprule
\textbf{Size} & \textbf{LUT (\%)} & \textbf{DFF (\%)} & \textbf{LUTRAM (\%)} & \textbf{DSP (\%)} & \textbf{BRAM (\%)} \\
\midrule
16   & 11.75 & 9.35  & 6.75  & 36.36 & 11.43 \\
32   & 21.94 & 13.52 & 2.29  & 72.73 & 28.57 \\
64   & 37.90 & 21.18 & 34.26 & 72.73 & 28.57 \\
128  & 54.16 & 30.75 & 67.54 & 72.73 & 34.29 \\
\bottomrule
\end{tabular}
\label{tab:fpga_resource_usage}
\end{table}

\subsubsection{MILP-Driven Optimal Platform-Task Configuration} 
Table~V illustrates the trade-offs between platforms and tasks, offering guidance for selecting optimal deployment strategies. FPGA (Platform 0) consistently achieves significantly lower runtime, making it well-suited for real-time, resource-constrained edge applications where low latency and energy efficiency are essential. For instance, ML on FPGA completes in just 20 seconds with moderate error, striking a strong balance between speed and performance. In contrast, GPU (Platform 1) excels in accuracy-critical scenarios—such as mission-critical MR tasks—achieving the lowest error (1.00) at the cost of substantially longer runtime (804.90 seconds). ML is preferable when rapid, approximate inference suffices, while MR is ideal when capturing interpretable, high-fidelity physical dynamics is the priority. In summary, FPGA + ML is best suited for fast and efficient monitoring at the edge, whereas GPU + MR is recommended for offline or high-accuracy applications demanding detailed physical insights. In practical deployments, users can flexibly choose the task-platform combination that best aligns with the specific performance, energy, and accuracy requirements of their application.

\begin{table}[thbp]
\centering
\scriptsize
\caption{MILP-selected hyperparameters and performance metrics across platforms and tasks.}
\renewcommand{\arraystretch}{1.2}
\begin{tabular}{ccccl}
\toprule
\textbf{Error} & \textbf{Time (s)} & \textbf{Platform} & \textbf{Task} & \textbf{Hyperparameters} \\
\midrule
5.00  & 20.00   & 0 & 0 & $e{=}16$, $N{=}50$, $hi{=}128$ \\
8.02  & 512.35  & 0 & 1 & $e{=}60$, $N{=}50$, $hi{=}16$  \\
5.40  & 446.37  & 0 & 2 & $e{=}128$, $N{=}200$, $hi{=}16$ \\
7.50  & 66.74   & 1 & 0 & $e{=}128$, $N{=}200$, $hi{=}16$ \\
19.16 & 174.25  & 1 & 1 & $e{=}16$, $N{=}50$, $hi{=}32$  \\
1.00  & 804.90  & 1 & 2 & $e{=}32$, $N{=}200$, $hi{=}16$ \\
\bottomrule
\end{tabular}
\label{tab:milp_result}
\end{table}

\section{Discussion}

The optimization framework in Fig.~\ref{fig:optimization_framework} illustrates a general principle for enabling efficient edge AI: replace computationally expensive iterative components with hardware-friendly, parallelizable alternatives while preserving functional equivalence. This principle extends beyond physical AI to any domain where energy consumption and inference latency are critical bottlenecks.

The framework naturally generalizes to Large Language Models (LLMs), which face similar challenges in edge deployment. Just as MERINDA replaces iterative ODE solvers with parallelizable GRU structures, analogous hardware-aware transformations can be applied to LLM architectures: (1) replacing autoregressive attention with linearized or state-space formulations amenable to FPGA pipelining, (2) substituting purely sequential decoding with speculative or partially parallel decoding schemes, and (3) exploiting FPGA fine-grained parallelism for sparse or structured attention patterns.

The significance of this generalization is substantial. Recent measurement studies show that state-of-the-art LLMs incur substantially higher energy cost per inference than traditional ML models---often an order of magnitude or more in comparable settings---due to their much larger parameter counts and GPU-intensive execution~\cite{Samsi2023FromWordsToWatts,Fernandez2025EnergyLLM,Ji2026ElectricityLLMs}. The 114$\times$ energy reduction and 2.96$\times$ latency improvement we observe for MERINDA on the model-recovery task demonstrate that hardware--algorithm co-design can deliver order-of-magnitude efficiency gains. While the exact factors will differ for language workloads, these results suggest that similar techniques could substantially reduce the energy and latency of LLM inference across a range of model scales---from compact edge models (1--7B parameters) to larger architectures. Such improvements would enable on-device language AI with a much lower carbon footprint and near-real-time responsiveness.


\section{Conclusions}
In this paper, we presented MERINDA, a novel FPGA-accelerated framework for model recovery (MR), designed to support physics-informed learning in mission-critical autonomous systems. Through both theoretical analysis and empirical validation, we demonstrated MERINDA’s equivalence to existing PiNODE architectures while achieving significant improvements in runtime, energy efficiency, and memory usage on FPGA. Our evaluations revealed clear trade-offs: FPGA is optimal for fast, energy-efficient deployment of model learning (ML) at the edge, while GPU excels in high-accuracy MR tasks suited for offline or high-performance environments. By exploring mixed-integer programming for task-platform selection, we showed how users can flexibly configure system behavior to meet varying constraints. Ultimately, our results highlight that strategic selection of task and hardware—guided by latency, power, and accuracy requirements—enables adaptive and efficient deployment of AI in resource-constrained, real-time applications.

\bibliographystyle{abbrv}
\bibliography{ref}

\end{document}